\documentclass[letterpaper, 10 pt, conference]{ieeeconf}  

\IEEEoverridecommandlockouts                              

\usepackage{epsfig} 
\usepackage{times} 
\usepackage{amsmath} 
\usepackage{amssymb}  
\usepackage{color}
\usepackage{xcolor}
\usepackage{preambles}
\usepackage{amsmath}
\usepackage{graphicx}
\usepackage{url} 
\usepackage[format=plain,labelsep=period]{caption}
\usepackage{wrapfig}
\usepackage[list=on,listformat=simple]{subcaption}
\usepackage{hyperref}
\usepackage{url}
\usepackage{caption}
\usepackage{subcaption}
\hypersetup{colorlinks = true, citecolor=black}

\title{\LARGE \bf
Real-Time Navigation for Bipedal Robots in Dynamic Environments}
\author{Octavian A. Donca$^{1}$, Chayapol Beokhaimook$^{2}$, and Ayonga Hereid$^{1}$
\thanks{*This work was supported in part by the National Science Foundation under grant FRR-21441568. }%
\thanks{$^{1}$Mechanical and Aerospace Engineering, Ohio State University, Columbus, OH, USA. {\tt\footnotesize (donca.2, hereid.1)@osu.edu.}}%
\thanks{$^{2}$Ottonomy Inc. {\tt\footnotesize chayapol.beokhaimook@gmail.com}}%
}

\usepackage[capitalise]{cleveref}

\usepackage[noadjust]{cite}

\begin{document}

\maketitle
\thispagestyle{empty}
\pagestyle{empty}

\begin{abstract}
The popularity of mobile robots has been steadily growing, with these robots being increasingly utilized to execute tasks previously completed by human workers. For bipedal robots to see this same success, robust autonomous navigation systems need to be developed that can execute in real-time and respond to dynamic environments. These systems can be divided into three stages: perception, planning, and control. A holistic navigation framework for bipedal robots must successfully integrate all three components of the autonomous navigation problem to enable robust real-world navigation. In this paper, we present a real-time navigation framework for bipedal robots in dynamic environments. The proposed system addresses all components of the navigation problem: We introduce a depth-based perception system for obstacle detection, mapping, and localization. A two-stage planner is developed to generate collision-free trajectories robust to unknown and dynamic environments. And execute trajectories on the Digit bipedal robot's walking gait controller. The navigation framework is validated through a series of simulation and hardware experiments that contain unknown environments and dynamic obstacles.
\end{abstract}

\section{Introduction}
\label{sec:intro}

Bipedal robots have been a popular field of research, due to the large range of tasks they can be utilized in. The general humanoid shape of bipedal robots allows them to be better integrated into a society designed for humans. However, for bipedal robots to be fully integrated into society, robust autonomous navigation systems need to be designed. These systems can generally be divided into three stages: perception, planning, and control. It is only with the combination of these three stages into a single navigation framework in which bipedal robots can truly integrate into human society. 

In this paper, we seek to implement a holistic bipedal robot navigation framework to enable the exploration of unknown, dynamic environments. The three components of autonomous navigation – perception, planning, and control – must be combined into a single navigation system. The perception system must be capable of extracting obstacle and environment structure information. This environment information must then be used to generate maps of the global and local environment for planning. A planning framework must generate global paths and local trajectories that are robust to unknown environments and dynamic obstacles. Furthermore, planning must respect the kinematic constraints of the robot while avoiding obstacles to ensure safe and feasible navigation. Finally, these trajectories must be executed with a low-level controller that maintains safe and stable walking gaits. The combination of these capabilities will enable the safe navigation of bipedal robots in complex environments.

Many works have expanded on the methods of A* \cite{chestnutt2007adaptive, huang2013energy, hornung2012anytime, garimort2011humanoid, 8460561}, PRM \cite{perrin2011weakly}, and RRT \cite{dalibard2013dynamic, baudouin2011real, nishi2014motion} to the unique problems of bipedal motion planning and footstep planning. However, many of these works lack several components required for autonomous navigation systems such as real-time perception, mapping, and localization processes. Furthermore, only few works expand further to adapt these bipedal motion planning methods into more holistic bipedal navigation frameworks. However, many are still unable to address components required in holistic autonomous bipedal systems such as on-robot perception systems \cite{michel2005vision}, localization methods \cite{7940036}, robustness to dynamic obstacles \cite{nishiwaki2012autonomous, 2109.05714}, or validation in hardware.

\begin{figure*}
    \centering
    \includegraphics[width=1.0\textwidth]{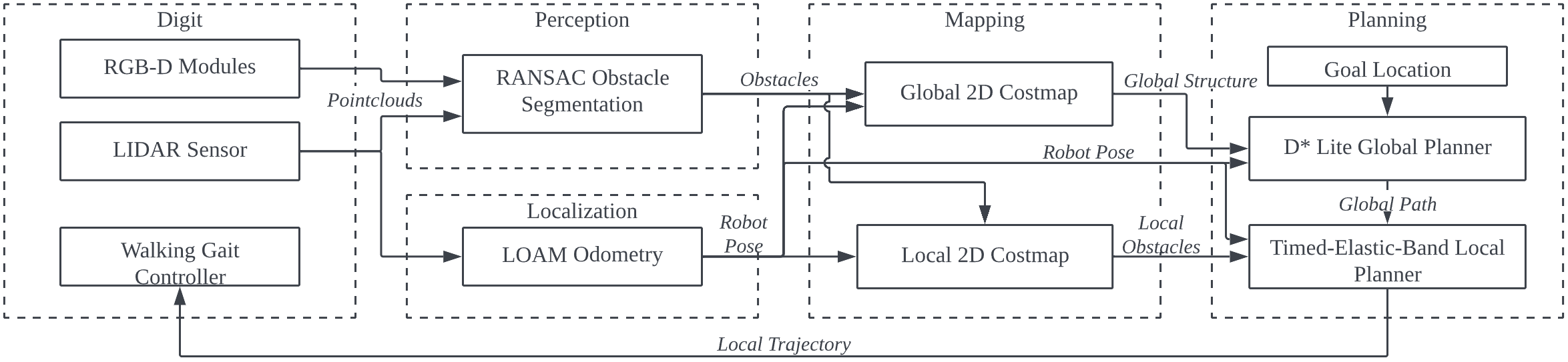}
    \setlength{\abovecaptionskip}{-10pt}
    \setlength{\belowcaptionskip}{-10pt}
    \caption{The proposed navigation framework, an architecture built on top of Move Base Flex \cite{8593829}. Point cloud detections are used for obstacle segmentation by Random Sample Consensus \cite{Fischler_Bolles_1981}. Global and local costmaps \cite{6942636} are generated from the obstacle segmentations. LiDAR Odometry and Mapping \cite{Zhang-2017-110808} localizes the robot. A D* Lite global planner \cite{d*lite} uses the global costmap to generate an optimal, collision-free path, which is used by the Timed-Elastic-Band local planner \cite{ROSMANN2017142} to generate local obstacle-avoiding trajectories, which are executed through velocity commands sent to Digit's gait controller.}
    \label{fig:system_architecture_overview}
\end{figure*}

We propose a real-time navigation framework for the Digit robot based on Move Base Flex~\cite{8593829}, as shown in \figref{fig:system_architecture_overview}. 
The framework utilizes two RGB-Depth cameras and a LiDAR sensor for perception. The environment is mapped using global and local costmaps to capture large-scale environment structure and local dynamic obstacles. Odometry and localization are calculated using LiDAR Odometry and Mapping during navigation. We developed a two-stage planner to generate collision-free paths and obstacle avoiding trajectories. In particular, a D* Lite global planner capable of fast re-planning is used to generate high-level paths in the global costmap. A Timed-Elastic-Band local planner then follows the global path through optimization-based trajectory generation that respects kinematic, velocity, acceleration, and obstacle avoidance constraints. The local trajectory is executed by generating a sequence of velocity commands sent to Digit's walking controller.

The rest of this paper is organized as follows. \secref{sec:percpetion_and_mapping} introduces the perception, mapping, and localization process. Next, \secref{sec:motion_planning} describes the motion planning methods. \secref{sec:results} introduces the simulation and hardware experiments and results. Finally, \secref{sec:conclusion} concludes the navigation framework and provides future work discussion.
\section{Digit Perception, Mapping, and Localization}\label{sec:percpetion_and_mapping}
In this section, we describe the process of building a real-time map of the environment using Digit's perception sensor suite and localizing Digit within that map. The robot is equipped with two depth cameras (Intel RealSense D430, placed at the pelvis, with one facing forwards at a downward angle and one facing rearwards at a downward angle), and one LiDAR sensor (Velodyne LiDAR VLP-16, placed on top of Digit’s torso).

\label{sec:architecture}

\subsection{Perception}
\begin{figure}
    \centering
    \includegraphics[width=1.0\linewidth]{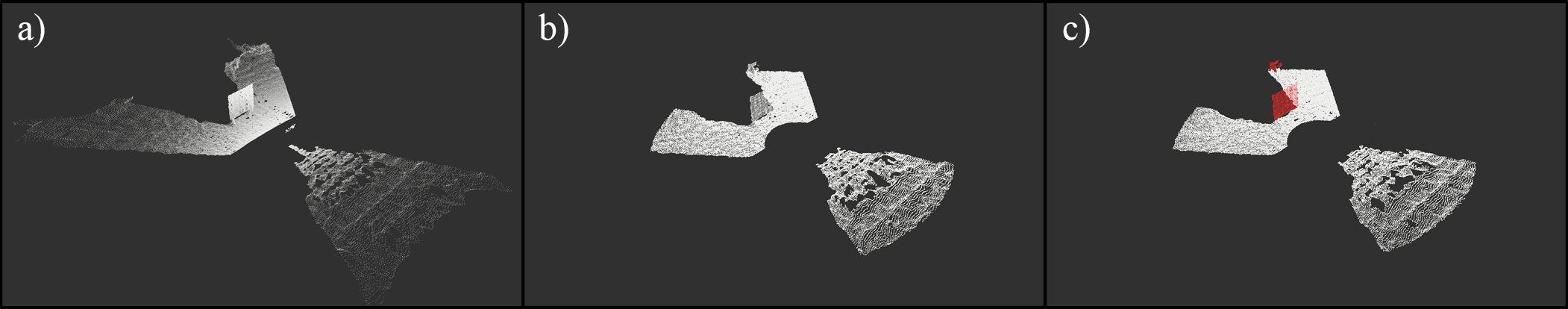}
    \setlength{\abovecaptionskip}{-10pt}
    \setlength{\belowcaptionskip}{-10pt}
    \caption{Pre-processing and obstacle segmentation results. a) Original point cloud, b) Filtered cloud, and c) Filtered point cloud with segmented obstacles in red.}
    \label{fig:obstacle_segmentation_digit}
\end{figure}
\newsec{Point Cloud Pre-processing.}
Before these point clouds can be used for obstacle segmentation, pre-processing is applied to obtain a uniform density and remove outlier detections. First, an average point cloud size reduction of 91.07\% is achieved using a Voxel Grid filter \cite{5980567} for downsampling. Additionally, inaccurate detections outside of the sensors' accurate range are removed by removing points further than 2.9 m of the depth cameras. Finally, erroneous points detected underground due to reflections are removed using a pass-through filter.

\newsec{Obstacle Segmentation.}
After filtering the point cloud from both depth cameras, the resulting clouds are fused into a combined point cloud. The Random Sample Consensus (RANSAC) method proposed in \cite{Fischler_Bolles_1981} is used to segment obstacles from the fused point cloud. For a given point cloud, $P$, RANSAC randomly samples 3 points to solve a unique plane model:
\begin{equation}
    ax + by + cz + d = 0,
\end{equation}
where \({a,b,c,d}\in\mathbb{R}\) are the fitted coefficients, and $(x,y,z)\in\mathbb{R}^3$ represents the Cartesian coordinates of a point.
Then, the absolute distance, \(D_i\), of each point \(i\) in the cloud is calculated for the fitted model:
\begin{equation}
    D_i=\left|\frac{a x_i+b y_i+c z_i+d}{\sqrt{a^2+b^2+c^2}}\right|, \text { for } i \in\{1,2, . ., n\},
\end{equation}
where $n$, is the total number of points in the point cloud $P$.
Points within a given threshold distance, \(D_{threshold}\), to the plane model are labeled as inliers of the model, denoted \(P_I\),
\begin{equation}
    P_I=\left\{p_i \in P \mid D_i<D_{\text {threshold}}\right\}.
\end{equation}
This process is repeated iteratively for a specified number of iterations, \(N\), determined statistically as: 
\begin{equation}
    N=\mathrm{round}\left(\frac{\log{(1-\alpha)}}{\log{(1-(1-\epsilon)^3)}}\right),
\end{equation}
where $\alpha$ is the desired minimum probability of finding at least one good plane from $P$, usually within \([0.90-0.99]\), and \(\epsilon=(1-u)\) where $u$ is the probability that any selected point is an inlier.
The resulting plane model is selected as the one which generates the largest number of inliers with the smallest standard deviation of distances.
From the resulting RANSAC plane model, inlier points are retained as ground points, and outlier points are retained as obstacle points.

\subsection{Mapping}
\label{sec:mapping}
To enable real-time planning in a dynamic environment, we create two 2D layered costmaps presented in \cite{6942636} to represent the global and local environments.

\subsubsection{Global Map}
The global map is used to capture and store macro-scale information of the environment. To enable this, the global map uses a lower resolution of 0.1 m and only updates at a rate of 2 Hz. Cells in the global map have two possible states:
\paragraph{Occupied}
Any 3D point detections from the perception sensors are projected onto the 2D plane of the global map. Any cells that contain projected points are classified as occupied and are considered untraversable.
\paragraph{Free}
Any cell that is not occupied is considered free and traversable for the robot. Unexplored regions of the environment are by default considered to be free.

\subsubsection{Local Map}
The local map is used for obstacle avoidance to capture a smaller region local to the robot where dynamic obstacles may be present. We use a higher resolution of 0.05 m and update the local map at 10 Hz. The local map introduces another cell state in addition to occupied and free cells:
\setcounter{paragraph}{2}
\paragraph{Inflated}
Cells within a certain distance of obstacles are inflated with non-zero cell costs. These non-zero costs are used to penalize trajectory planning through regions close to obstacles while not completely prohibiting trajectories from entering these regions.

\subsection{Localization}
In this work, we use LiDAR Odometry and Mapping (LOAM), a method introduced in \cite{Zhang-2017-110808}, to estimate the odometry of the robot. LOAM introduces simultaneously executed algorithms to estimate LiDAR motion: an odometry estimation algorithm and a mapping algorithm. 

The LiDAR odometry algorithm runs at 10 Hz in which the correspondences of extracted features from consecutive LiDAR sensor sweeps are used to estimate the motion of the sensor. For each sensor sweep, LOAM extracts points from edge line and planar patch environment structures as features. These edge lines and planar patches in the environment are referenced as correspondences for feature points extracted from the respective environment structures.

Using the correspondences, geometric relationships can be formed between an edge point and its corresponding edge line and a planar point and its corresponding planar patch:
\begin{align}
    f_{\mathcal{E}}\left(\boldsymbol{X}_{(k+1, i)}^L, \boldsymbol{T}_{k+1}^L\right) &= d_{\mathcal{E}}, i \in \mathcal{E}_{k+1},\\
    f_{\mathcal{H}}\left(\boldsymbol{X}_{(k+1, i)}^L, \boldsymbol{T}_{k+1}^L\right) &= d_{\mathcal{H}}, i \in \mathcal{H}_{k+1},
\end{align}
where \(\boldsymbol{T}_{k+1}^L\) is the LiDAR pose transform from \([t_{k+1}, t]\), $\mathcal{E}_{k+1}$ and $\mathcal{H}_{k+1}$ are the sets of edge points and planar points from the point cloud at $t_{k+1}$, and $d_{\mathcal{E}}$ and $d_{\mathcal{H}}$ are the distances from the corresponding edge and planar feature points to their respective edge line and planar patch environment structures.
These relationships are stacked for each edge and planar feature point to obtain a nonlinear function,
\begin{equation}
    \boldsymbol{f}\left(\boldsymbol{T}_{k+1}^L\right)=\boldsymbol{d},
\end{equation}
where each row of $\boldsymbol{f}$ is a feature point and each row of $\boldsymbol{d}$ is the correspondance distance.
This function is solved with the Levenberg-Marquardt method \cite{Hartley:2003:MVG:861369} to obtain the motion estimation of the LiDAR sensor. 

\begin{figure}
\vspace{2mm}
    \centering
    \includegraphics[width=1.0\linewidth]{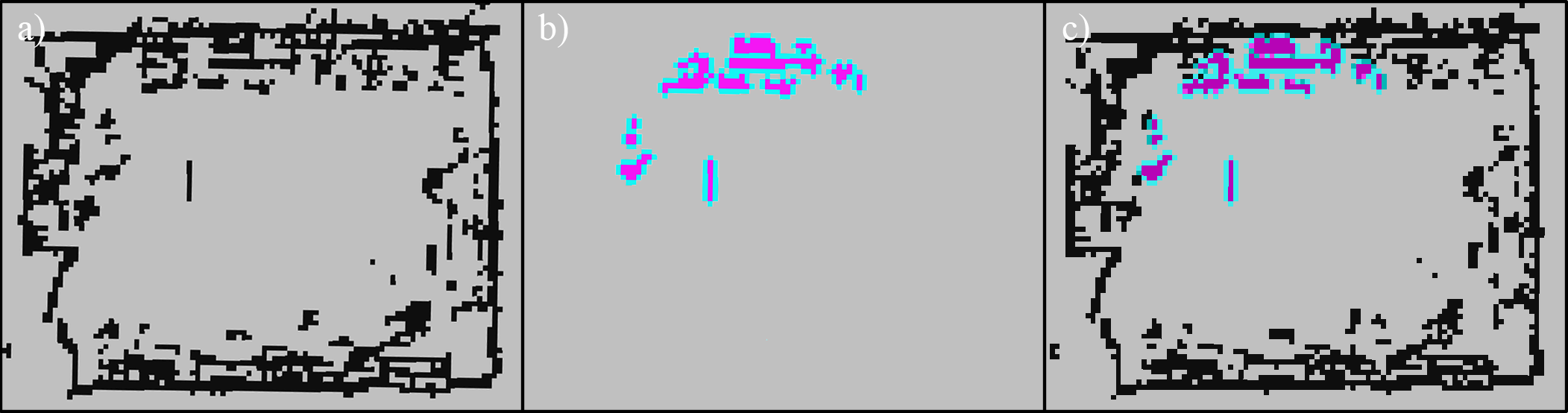}
    \setlength{\abovecaptionskip}{-10pt}
    \setlength{\belowcaptionskip}{-10pt}
    \caption{Mapping results from the same environment as shown in \figref{fig:obstacle_segmentation_digit}. a) Global map. b) Local map. c) Overlaid global and local maps.}
    \label{fig:digit_mapping_example}
\end{figure}

The final result from this localization method is an estimate of the LiDAR sensor odometry relative to the 3D environment. The coordinate frame origin of this odometry is initialized to the initial position of the LiDAR sensor. The mapping methods mentioned in \secref{sec:mapping} initialize the global origin as the 2D components of this LOAM origin. We then use this transform to localize the LiDAR sensor, and thus the robot, in the 2D costmaps during navigation. 

\section{Real-Time Motion Planning}\label{sec:motion_planning}

Using the global and local map created through the aforementioned process, we developed a real-time motion planning framework for Digit using D* Lite and Timed-Elastic-Band. 

\subsection{D* Lite global planner}
During navigation through unknown environments, collision-free paths through the global environment are required. As unknown environments are explored, a planner must be capable of quick re-planning to account for discovered obstacles. Therefore, a D* Lite global planner as presented in \cite{d*lite} is used to generate these global paths. Capable of fast re-planning, D* Lite is designed to solve goal-directed navigation problems in grid-represented environments. This planning method is directly applied to the 2D layered costmaps described in \secref{sec:mapping}.

The D* Lite planner assigns each explored cell, \(s\), in the costmap two values:
an estimate of the goal distance, \(g(s)=h(s, s_{start})\), where \(h\) is the Euclidean distance from the start cell \(s_{start}\) to the current cell \(s\); and a one-step lookahead value based on the g-value, \(rhs(s)\):
\begin{equation}
    rhs(s)= \begin{cases}0 &\text {if } s=s_{s t a r t} \\ \min _{s^{\prime} \in \operatorname{Pred}(s)}\left(g\left(s^{\prime}\right)+c\left(s^{\prime}, s\right)\right) &\text {otherwise}\end{cases}
\end{equation}
where \(c(s', s)\) is the cost to traverse to a neighboring cell, \(s'\), from the current cell, \(s\).

To generate the shortest path, the planner begins from the goal position and explores towards the start position. All neighbors of the current cell are added to a priority queue with a key value \(k(s) = [k_1(s), k_2(s)]\) with
\begin{align}
    k_1(s) &= \min(g(s), rhs(s)) + h(s, s_{start}) + k_m,\\
    k_2(s) &= \min(g(s), rhs(s)),
\end{align}
where $k_m$ is a scalar whose value is incremented each time the planner re-plans, and is used to maintain the priority queue ordering.
The priority queue is sorted by the following sort rule: a key \(k(s) \leq k'(s)\) iff \(k_1(s) < k'_1(s)\) or \(k_1(s) = k'_1(s)\) and \(k_2(s) < k'_2(s)\).
The planner always expands the lowest-valued cell in the priority queue. Then, the successor cells, or neighbors, of this expanded cell have their \(g\) and \(rhs\) values updated and are then themselves inserted into the priority queue. This process ends once the start cell is expanded or no more cells can be expanded. 

The main planning algorithm initially generates the shortest path based on the initial environment information. As the robot explores new environment areas and obstacles are detected, the costmap, and corresponding $c$ values, is updated to reflect the new obstacle detections and the global planner re-calculates the global path as previously described. The current path is stored as a sequence of waypoint cells within the global costmap to follow and is updated at 5 Hz. 

\subsection{Timed-Elastic-Band local planner}
The purpose of the local planner is to generate a smooth, kinematically feasible trajectory that follows the global path generated by the D* Lite global planner. The Timed-Elastic-Band (TEB) method introduced in \cite{ROSMANN2017142} originally implements a trajectory optimization method for differential models. We adapt this method for use in bipedal robots by applying simplifying assumptions on the high-level kinematic constraints of the low-level controller to mimic those of a differential drive robot. The bipedal locomotion is modeled as a differential drive model to avoid several problems introduced from the omnidirectional movement of bipedal robots. For example, lateral velocity in bipedal locomotion is difficult to control due to oscillating behaviors in the lateral direction. 

The TEB problem is defined as an open-loop optimization task to find the sequence of controls needed to move the robot from an initial pose, \(\mathbf{s}_s\), to a goal pose, \(\mathbf{s}_f\). A trajectory is defined as a sequence of robot poses \(\mathcal{S} = \{\mathbf{s}_k | k = 1, 2, …, n\}\) where \(\mathbf{s}_k =[x_k, y_k, \beta_k]^{\top}\) denotes the robot pose at time \(k\), with \(x_k\) and \(y_k\) denoting the 2D position of the robot and \(\beta_k\) denoting the orientation of the robot.  The TEB planner then augments the trajectory with positive time intervals \(\Delta T_k, k = 1, 2, …, n\). The sequence of robot poses and time intervals are then joined to form the parameter vector \(\mathbf{b}=\left[\mathbf{s}_1, \Delta T_1, \mathbf{s}_2, \Delta T_2, \mathbf{s}_3, \ldots, \Delta T_{n-1}, \mathbf{s}_n\right]^{\top}\). TEB solves the non-linear optimization task defined as:
\begin{equation}\label{eq:loam_nonlinear_task}
    V^*(\mathbf{b})=\min_{\mathbf{b}}{\sum_{k=1}^{n-1} \Delta T_k^2},
\end{equation}
which is subject to several constraints defined as equality and inequality constraints.

\paragraph{Non-holonomic constraint}
An equality constraint \(\mathbf{v}_k\) enforces the kinematic constraints of the robot. This non-holonomic constraint can be interpreted geometrically to assume consecutive poses \(\mathbf{s}_k\) and \(\mathbf{s}_{k+1}\) must share a common arc of constant curvature \cite{Rsmann2012TrajectoryMC}. The angle \(\mathbf{\vartheta}_k\) between pose \(\mathbf{s}_k\) and direction \(\mathbf{d}_{k, k+1} = [x_{k+1}-x_k, y_{k+1}–y_k, 0]^{\top}\) must be equal to angle \(\vartheta_{k+1}\) at pose \(\mathbf{s}_{k+1}\), with \(\vartheta_k = \vartheta_{k+1}\) which can be replaced with:
\begin{equation}
    \begin{bmatrix}
        \cos (\beta_k) \\
        \sin (\beta_k) \\
        0
    \end{bmatrix} \times \mathbf{d}_{k, k+1} = \mathbf{d}_{k, k+1} \times \begin{bmatrix}
        \cos (\beta_{k+1}) \\
        \sin (\beta_{k+1}) \\
        0
    \end{bmatrix},
\end{equation}
and thus the resulting equality constraint applied to the optimization task in \eqref{eq:loam_nonlinear_task} is given by:
\begin{equation}
    \mathbf{h}_k(\mathbf{s}_{k+1}, \mathbf{s}_k) = \begin{pmatrix}
        \begin{bmatrix}
            \cos (\beta_k) \\
            \sin (\beta_k) \\
            0
        \end{bmatrix} + \begin{bmatrix}
            \cos (\beta_{k+1}) \\
            \sin (\beta_{k+1}) \\
            0
        \end{bmatrix}
    \end{pmatrix} \times \mathbf{d}_{k, k+1}. 
\end{equation}
\paragraph{Velocity and acceleration constraints}
Additionally, limitations of the linear and angular velocities and accelerations are applied. First, the linear and angular velocities are approximated between two consecutive poses \(s_k\) and \(s_{k+1}\):
\begin{align}
    v_k&=\Delta T_k^{-1}\lVert [x_{k+1}-x_k, y_{k+1}-y_k]^{\top} \rVert\gamma(\mathbf{s}_k,\mathbf{s}_{k+1}),\\
    \omega_k&=\Delta T_k^{-1}(\beta_{k+1}-\beta_k),
\end{align}
where \(\gamma(\mathbf{s}_k,\mathbf{s}_{k+1})\) is a sign extraction function which is approximated by a smooth sigmoidal approximation that maps to the interval \([-1, 1]\):
\begin{equation}
    \gamma(\mathbf{s}_k,\mathbf{s}_{k+1}) \approx \frac{\kappa \langle \mathbf{q}_k, \mathbf{d}_{k, k+1} \rangle}{1 + | \kappa \langle \mathbf{q}_k, \mathbf{d}_{k, k+1} \rangle |}.
\end{equation}
Next, the accelerations are approximated in a similar manner to the velocities between consecutive poses:
\begin{equation}
    a_k=\frac{2(v_{k+1}-v_k)}{\Delta T_k + \Delta T_{k+1}}; \dot{\omega}_k=\frac{2(\omega_{k+1}-\omega_k)}{\Delta T_k + \Delta T_{k+1}}.
\end{equation}
Finally, the velocity and acceleration constraints are applied to \eqref{eq:loam_nonlinear_task} as:
\begin{align}
    \mathbf{v}_k(\mathbf{s}_{k+1}, \mathbf{s}_{k}, \Delta T_k) &=\left[
    \begin{array}{c}
         v_{max}-|v_k| \\
         \omega_{max}-|\omega_k| 
    \end{array}\right],\nonumber \\
    \boldsymbol{\alpha}_k(\mathbf{s}_{k+2}, \mathbf{s}_{k+1}, \mathbf{s}_{k}, \Delta T_{k+1}, \Delta T_{k}) &= \left[
    \begin{array}{c}
         a_{max}-|a_k| \\
         \dot\omega_{max}-|\dot\omega_k|
    \end{array}\right].\nonumber
\end{align}
\begin{figure*}
    \centering
    \includegraphics[width=1.0\textwidth]{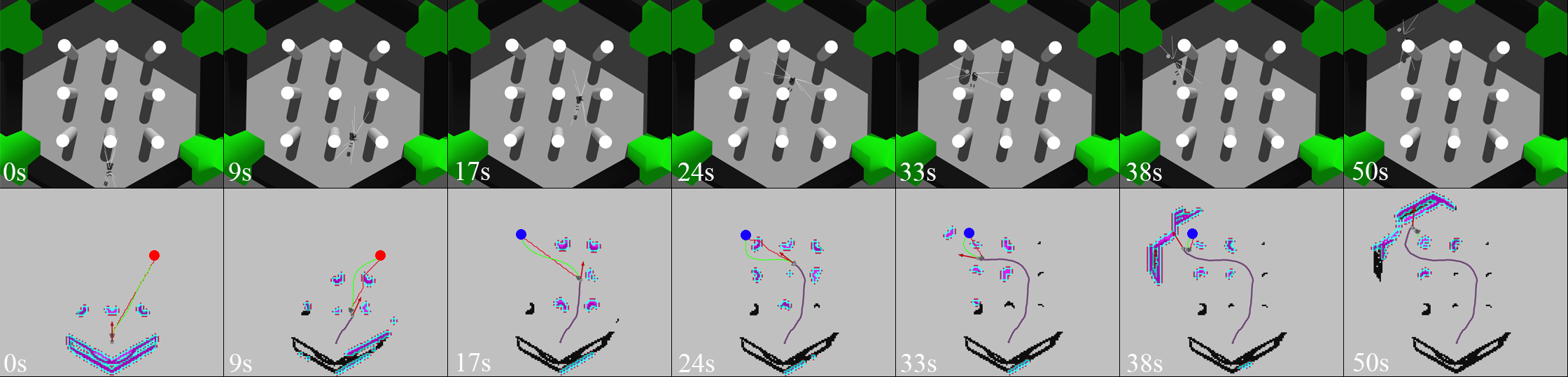}
    \setlength{\abovecaptionskip}{-10pt}
    \caption{Experiment \ref{sssec:sim_exp_1}) A differential drive robot navigates to a goal in a static environment. As unknown obstacles are discovered, re-planning occurs. Midway through execution, a new goal is provided and is reached.}
    \label{fig:sim_static_experiment_1}
\end{figure*}
\begin{figure*}
    \centering
    \includegraphics[width=1.0\textwidth]{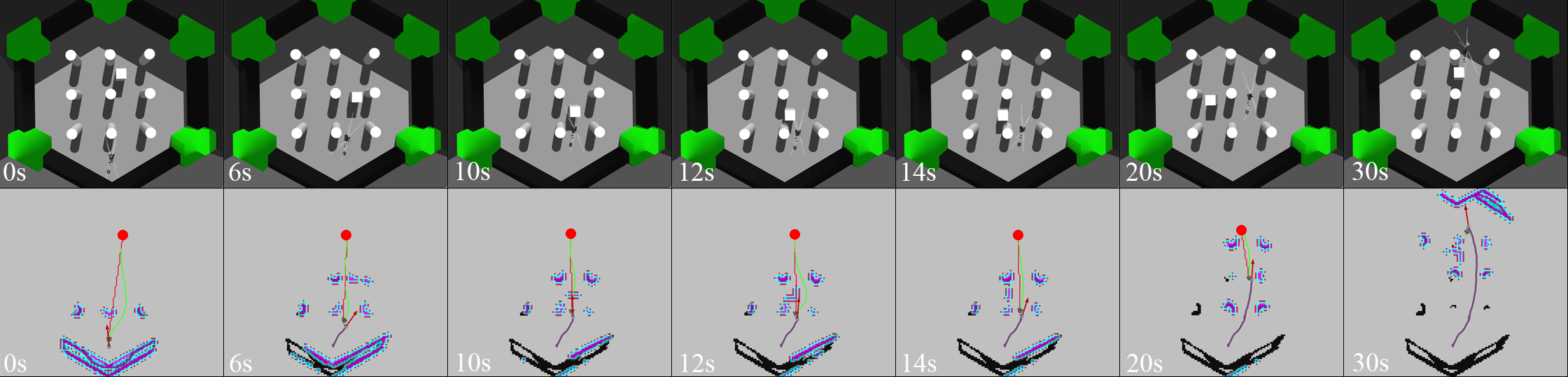}
    \setlength{\abovecaptionskip}{-10pt}
    \setlength{\belowcaptionskip}{-15pt}
    \caption{Experiment \ref{sssec:sim_exp_2}) A differential drive robot navigates to a goal in a dynamic environment. As the unknown dynamic obstacles is discovered, the robot stops, begins to reverse, and then plans around the obstacle. Once the dynamic obstacle has passed, the robot begins to navigate forward and successfully reaches the goal.}
    \label{fig:sim_static_experiment_2}
\end{figure*}
\paragraph{Obstacle avoidance}
The local planner considers obstacles, \(\mathcal{O}_l, l=1, 2, ..., R\), as simply-connected regions in the form of points, circles, polygons, and lines in \(\mathbb{R}^2\). To ensure a minimum separation \(\rho_{min}\) between pose \(\mathbf{s}_k\) and all obstacles, the inequality constraint applied to \eqref{eq:loam_nonlinear_task} is:
\begin{align}
    \begin{split}
        \mathbf{o}_k(\mathbf{s}_k)=[\rho(\mathbf{s}_k, \mathcal{O}_1), \rho(\mathbf{s}_k, \mathcal{O}_2), ..., \rho(\mathbf{s}_k, \mathcal{O}_R)]^{\top} \\
        - [\rho_{min}, \rho_{min}, ..., \rho_{min}]^{\top},
    \end{split}
\end{align}
where \(\rho\) is the minimal Euclidean distance between the obstacle and the robot pose.

The nonlinear task \eqref{eq:loam_nonlinear_task} is reformulated into an approximate nonlinear least-squares optimization problem for better computational efficiency. With the constraints applied to the approximated objective function as described in \cite{Nocedal_Wright}. The kinematic equality constraint $\mathbf{h}_k$ is expressed as:
\begin{equation}
    \phi(\mathbf{h}_k, \sigma_h)=\sigma_h\lVert\mathbf{h}_k\rVert_2^{2},
\end{equation}
and the inequality constraint $\mathbf{v}_k$ is approximated as:
\begin{equation}
    \chi(\mathbf{v}_k, \sigma_v)=\sigma_v\lVert\min\{\mathbf{0}, \mathbf{v}_k\}\rVert_2^{2},
\end{equation}
where \(\sigma_i, \sigma_v\) are scalar weights and the min operator is applied row-wise and with constraints $\boldsymbol{\alpha}_k$ and $\mathbf{o}_k$ being approximated in an identical manner.
The overall optimization problem is approximated with the objective function: 
\begin{align}\label{eq:loam_kin_constraint_approx_1}
    \mathbf{b}^*=\argmin_{\mathcal{B}\backslash\{\mathbf{s}_1,\mathbf{s}_n\}} \tilde{V}(\mathbf{b}),
    \vspace{-3mm}
\end{align}
\begin{align}
\hspace{-3mm}
\label{eq:loam_kin_constraint_approx_2}
        \tilde{V}(\mathbf{b})=\sum_{k=1}^{n-1}[&\Delta T_k^2 + \phi(\mathbf{h}_k, \sigma_h) + \chi(r_k, o_k) + \chi(\mathbf{v}_k, \sigma_v) \nonumber\\ 
        & + \chi(\boldsymbol{\alpha}_k, \sigma_{\alpha}) + \chi(\mathbf{o}_k, \sigma_o)] + \chi(\boldsymbol{\alpha}_n, \sigma_{\alpha}).
\end{align}
This optimization problem is solved using a variant of the Levenberg–Marquardt algorithm implemented in the open-source graph optimization framework g2o \cite{Kmmerle2011G2oAG}.
As described in \cite{ROSMANN2017142}, the Timed-Elastic-Band planner further improves the method by optimizing several of these trajectories from distinctive topologies in parallel to search for the global minimum of \eqref{eq:loam_nonlinear_task} respecting \eqref{eq:loam_kin_constraint_approx_1}.

The final output from the Timed-Elastic-Band planner is sequences of velocity commands to achieve the globally optimal trajectory \(\mathbf{b}^*\) from \eqref{eq:loam_kin_constraint_approx_1}. These velocity commands are sent to Digit’s default low-level controller provided by Agility Robotics.

\section{Results}
\label{sec:results}
This section presents the simulation results of the proposed approach and the experimental validation with Digit in dynamic and unknown environments\footnote{Experiment recordings are shown in the following video: \href{https://www.youtube.com/watch?v=WzHejHx-Kzs}{https://www.youtube.com/watch?v=WzHejHx-Kzs}.}.

\begin{figure*}
    \centering
    \includegraphics[width=1.0\textwidth]{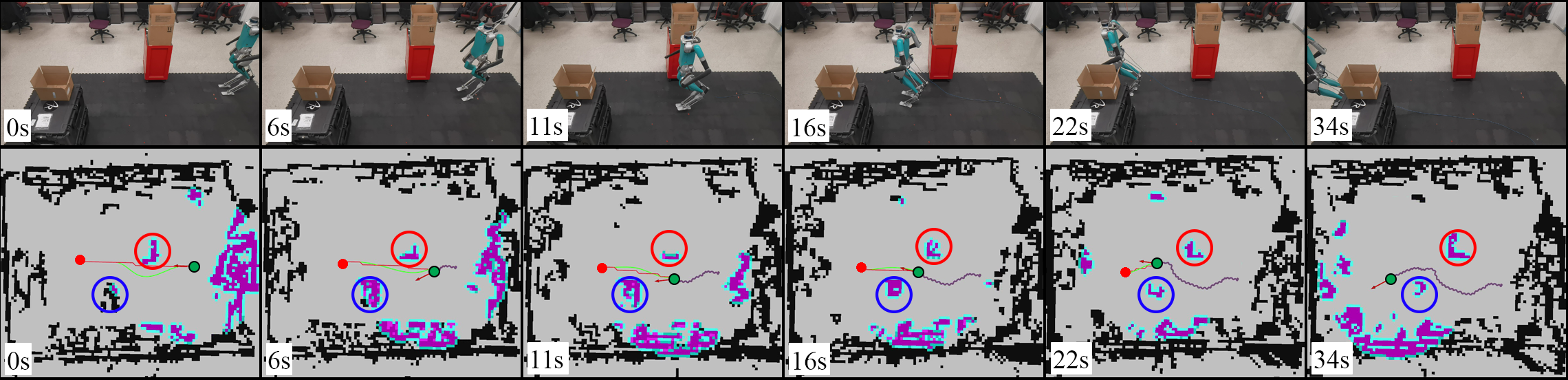}
    \setlength{\abovecaptionskip}{-10pt}
    \caption{Experiment (\ref{sssec:digit_exp_1}) Digit navigating to a goal in a static environment. Digit is able to successfully navigate around static obstacles blocking the initial trajectory and safely reaches the goal position.}
    \label{fig:digit_static_experiment_1}
\end{figure*}
\begin{figure*}
    \centering
    \includegraphics[width=1.0\textwidth]{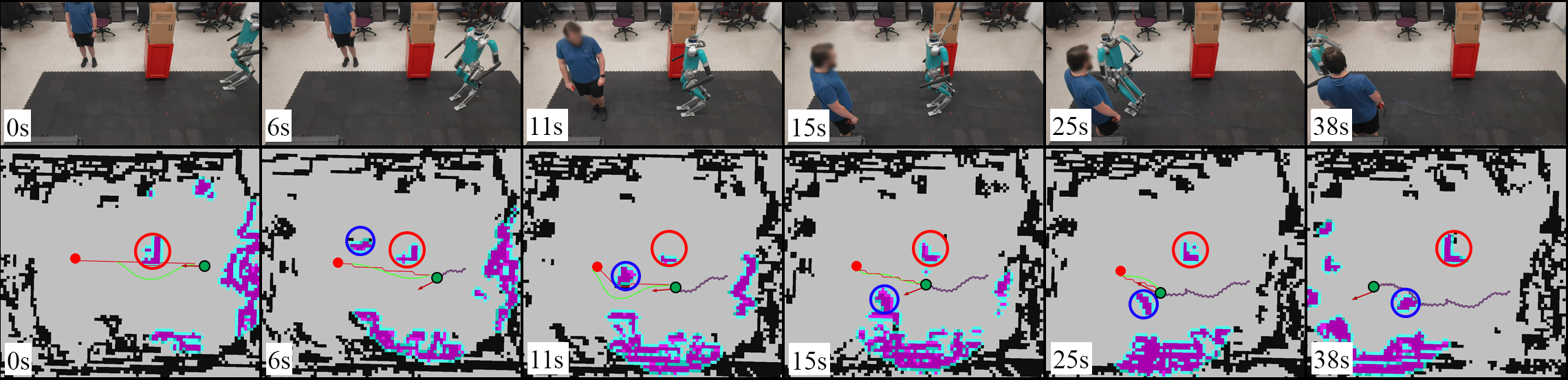}
    \setlength{\abovecaptionskip}{-10pt}
    \setlength{\belowcaptionskip}{-10pt}
    \caption{Experiment (\ref{sssec:digit_exp_2}) Digit navigating to a goal in a dynamic environment. During navigation, a dynamic obstacle is discovered and as the obstacle begins to move, the trajectory deforms to maintain a collision free path. The dynamic obstacle finally comes to a rest and Digit is able to successfully navigate around the obstacles and safely reach the goal position.}
    \label{fig:digit_dynamic_experiment_1}
\end{figure*}
\subsection{Simulation Experiments}
The framework is initially validated in simulation. In experiments shown in \figref{fig:sim_static_experiment_1} and \figref{fig:sim_static_experiment_2}, the red and blue circles represent the first and second goals. Black grid cells are obstacles in the global map, purple and blue grid cells are obstacles and inflated cells in the local map, respectively. The red arrow is the current estimated odometry. The red path and green paths are the global path and local trajectory, respectively. The purple path shows the robot's traveled path.
\subsubsection{Static, unknown environment}{\label{sssec:sim_exp_1}}
As shown in \figref{fig:sim_static_experiment_1}, a 6.0 m $\times$ 6.0 m simulated environment contains nine static, cylindrical obstacles with radius 0.15 m evenly spaced apart. At $t=0$, an initial goal position is selected. From $t=0$ to $t=9$ the robot navigates along the path until an obstacle is discovered obstructing the current trajectory, at which point the global path and local trajectory are quickly re-planned to account for the new obstacle. While continuing to navigate to the first goal, a new goal position is given. The framework quickly re-plans and begins navigating to the new goal. At $t=24$, a blocking obstacle is again detected and re-planning occurs. From $t=24$ to $t=50$ the robot successfully executes the trajectory and reaches the goal position.
\subsubsection{Dynamic, unknown environment}{\label{sssec:sim_exp_2}}
The second simulated experiment occurs in the same environment as the previous experiment, except a dynamic obstacle has been added. At $t=0$, an initial goal position is selected. The robot begins to navigate until $t=6$ when the dynamic obstacle comes into view and blocks the path. The planners attempt to re-plan around the obstacle but determine the current path is untraversable. The robot stops and begins to reverse until $t=12$ when a viable trajectory is discovered. At $t=14$, the robot stops reversing and begins to execute the trajectory as the dynamic obstacle has passed. From $t=14$ to $t=30$, the robot executes the trajectory and reaches the goal.
\subsection{Hardware Experiments}
The navigation framework computation is executed on an external desktop using an AMD Ryzen 5800X CPU and connected through a TCP connection to Digit.
In this work, two hardware experiments were conducted indoors: 1) Planning in a static environment and 2) Planning in a dynamic environment. Each test is deemed successful once the Digit robot is able to navigate collision-free from the start position to the goal position within a 0.2 m final goal position tolerance and 0.2 rad final goal orientation tolerance. In experiments shown in \figref{fig:digit_static_experiment_1} and \figref{fig:digit_dynamic_experiment_1} the solid red circle represents the goal position. The red and blue hollow circles represent the first and second obstacles, the second being a dynamic obstacle in \figref{fig:digit_dynamic_experiment_1}. Black grid cells are obstacles in the global map, purple and blue grid cells are obstacles and inflated cells in the local map, respectively. The red arrow is the current estimated odometry, and the green circle shows the robots current position. The red path and green paths are the global path and local trajectory, respectively. The purple path shows the robot's traveled path.
\subsubsection{Static environment}{\label{sssec:digit_exp_1}}
As shown in \figref{fig:digit_static_experiment_1}, a 10.0 m $\times$ 8.0 m environment contains two static 0.5 m $\times$ 1.0 m and 1.0 m $\times$ 0.5 m obstacles. At $t=0$ a goal position is given to Digit and the robot begins to rotate to safely avoid the first obstacle. Digit continues to travel forward between $t=0$ and $t=11$ at which point Digit begins to rotate to avoid the second obstacle. Digit continues along the generated trajectory until the robot safely reaches the goal position at $t=34$.
\subsubsection{Dynamic environment}{\label{sssec:digit_exp_2}}
The second experiment is conducted in the same environment as the previous experiment, except the second static obstacle has been replaced with a dynamic obstacle. At $t=0$ a goal position is given to Digit. Initially, the dynamic obstacle is not visible to Digit due to being obscured by the first obstacle. Digit begins to execute the generated trajectory and at $t=6$ the dynamic obstacle is detected. At this moment, the dynamic obstacle begins to navigate across the currently planned trajectory. As the obstacle moves in the environment, the planned trajectory begins to deform to maintain a collision-free path, which can be seen at $t=11$. Between $t=11$ and $t=15$, Digit begins to slow to a stop as the dynamic obstacle continues to move across the robots path. At $t=15$, the obstacle comes to a rest and Digit is able to begin executing a collision-free path to the goal. Digit is able to successfully execute the trajectory between $t=15$ and $t=38$ and safely reaches the goal position at $t=38$.
\section{Conclusions}
\label{sec:conclusion}

In this work, we have presented a real-time navigation framework for bipedal robots in dynamic environments. Successful experiments were conducted in both simulation and hardware.

While the proposed navigation framework was shown to successfully execute in complex environments, there exist several limitations. The current sensor suite configuration results in blind spots on the left and right sides of the robot. This prevents the navigation framework from quickly reacting to dynamic obstacles that move in these blind spots. In future work, a more robust sensor suite providing $360^o$ sensor coverage would allow for better dynamic obstacle reactions. 
Additionally, the use of 2D layered costmaps limits the navigable environments to only flat 2D terrains. While this is sufficient for many complex environments, it is often desirable to navigate through inclines, stairs, and uneven terrain. Using a mapping and planning framework that can utilize 3D information would allow for more robust navigation in complex 3D terrains. 
Finally, although the local planner is able to generate robust trajectories, the assumption of differential drive kinematic constraints limits the capabilities of bipedal robots. An extension to omnidirectional locomotion would allow for better utilization of the unique capabilities of bipedal robots.

\newpage
\bibliographystyle{IEEEtran}
\bibliography{references.bib}

\end{document}